%% file: main.tex
\documentclass[submission]{sacjsub}

\usepackage{graphicx, color} 
\usepackage[font={small}]{caption}
\usepackage{array, booktabs, longtable} 
\usepackage{amsmath, array, subcaption}
\usepackage{pdflscape, rotating}
\usepackage{hyperref}
\usepackage{microtype}
\usepackage{relsize}
\usepackage{multicol}
\begin{document}

\SACJtitle{Clustering Residential Electricity Consumption Data to Create Archetypes that Capture Household Behaviour in South Africa}

\SACJauthor[tu,uct]{Wiebke Toussaint}{w.toussaint@tudelft.nl}{corresponding}
\SACJauthor[uct,cair]{Deshendran Moodley}{deshen@cs.uct.ac.za}{}
 
\SACJaddress[tu]{Engineering Systems \& Services Department, Delft University of Technology, Netherlands}
\SACJaddress[uct]{Department of Computer Science, University of Cape Town, South Africa}
\SACJaddress[cair]{Centre for Artificial Intelligence Research, South Africa}
 
\SACJrunningheader{Toussaint, W ~and Moodley, D.}{Clustering Residential Electricity Consumption Data to Create Archetypes}

\SACJcitationauthors{Toussaint, W ~and Moodley, D.}

\SACJabstract{
Clustering is frequently used in the energy domain to identify dominant electricity consumption patterns of households, which can be used to construct customer archetypes for long term energy planning. Selecting a useful set of clusters however requires extensive experimentation and domain knowledge. While internal clustering validation measures are well established in the electricity domain, they are limited for selecting useful clusters. Based on an application case study in South Africa, we present an approach for formalising implicit expert knowledge as external evaluation measures to create customer archetypes that capture variability in residential electricity consumption behaviour. By combining internal and external validation measures in a structured manner, we were able to evaluate clustering structures based on the utility they present for our application. We validate the selected clusters in a use case where we successfully reconstruct customer archetypes previously developed by experts. Our approach shows promise for transparent and repeatable cluster ranking and selection by data scientists, even if they have limited domain knowledge.
}

\SACJACMCategory{Computing methodologies}{Cluster analysis}{h}
\SACJACMCategory{Applied computing}{Engineering}{h}
\SACJkeywords{clustering, external clustering validation measures, competency questions, daily household electricity patterns, customer segmentation}

\SACJmaketitle

\input{sections/1_introduction}
\input{sections/2_literature}
\input{sections/3_clustering}
\input{sections/4_application_requirements}
\input{sections/5_results}
\input{sections/6_archetypes}
\input{sections/7_discussion}
\input{sections/8_conclusion}

\bibliographystyle{unsrtnat}
\bibliography{library}

\appendix
\include{sections/appendix_b}
\include{sections/appendix_c}

\end{document}

%% file: sections/1_introduction.tex
\section{Introduction}
\label{introduction}

Energy planning requires insights into the electricity consumption behaviour of customers to predict long term demand. Unlike commercial and industrial customers who consume electricity predictably, the daily consumption behaviour of residential households is highly variable \citep{Swan2009}. In South Africa economic volatility, income inequality, geographic and social diversity contribute to increased variability of daily household consumption behaviour~\citep{Heunis2014}. Understanding this variability is important for policy and planning decisions such as tariff design, network planning and operation, and demand response programmes \citep{Yilmaz2019}. 

The aggregate consumption behaviour, or representative load profiles, of residential customers has been modelled extensively to yield standard consumption patterns, or \textit{archetypes}, for dominant groups of households that have common attributes \cite{Swan2009}. These archetypes consolidate expert knowledge and represent the electricity consumption of typical customer classes. They are an essential tool for demand planning, but are difficult and tedious to construct and do not cater for changes in household behaviour. This is a serious limitation that impacts energy demand planning. Daily consumption behaviour can vary drastically for individual households over time~\cite{Dent2014a}. In addition, several years may pass before archetypes are updated. The dominant customer classes can become outdated as new groups of households emerge which may not correspond to the current archetypes. An example of this is households in rural areas in South Africa, where thatch roof huts with limited appliances have gradually transitioned to brick buildings with modern appliances, resulting in a significant change in electricity consumption. 


Cluster analysis can be used to identify dominant daily energy consumption patterns for different types of households. Current approaches usually aggregate households~\cite{Dang-Ha2017}\cite{McLoughlin2015} or assume that their consumption behaviour is static. This limits them in their ability to create archetypes that consider variability in daily consumption patterns over time. Another challenge is cluster validation. As acknowledged in the data mining community, clustering problems are notoriously difficult to evaluate. Internal validation measures are frequently limited to specific application scenarios, and insufficient on their own \cite{Liu2010}. External evaluation measures can be used instead, but they require ground-truth labels that indicate true clusters \cite{Song2008}. These are often not available because clustering is an unsupervised learning problem and true clusters are typically unknown. Evaluation by visual inspection is thus relied on, but can be biased by the interpretation of the visual representation \citep{Gogolou2019}. In the electricity domain these challenges are evident. While internal metrics are well established, cluster rankings produced by internal metrics yield conflicting results and are usually insufficient for discriminating between different cluster sets~\citep{Jin2017}. Cluster selection is thus typically done through visual inspection, which can be time consuming, adhoc, subjective and difficult to reproduce. When archetypes are updated, the visual evaluation process also has to be repeated. External validation measures based on domain knowledge are sometimes used to rank and guide the selection of a useful cluster set \cite{Xu2017}. However, there are no standard external metrics or guidelines for choosing such measures.

The data mining literature suggests that cluster quality is best evaluated against the specific objectives of the application \cite{Aggarwal2015}. Even so, data scientists often have limited domain knowledge, which can hinder them from identifying useful clustering structures. The field of ontology engineering provides structured methods for acquiring and representing knowledge from domain experts. One such method, competency questions, is widely used by ontology engineers to acquire application requirements and to compare and evaluate candidate conceptualisations of domain knowledge for a specified context~\cite{Gruninger1995}. Using an application case study in South Africa, we present an approach that uses informal interviews to derive competency questions, which we operationalise as external evaluation measures to identify a cluster set that represents the most useful daily consumption patterns in our dataset for analysing variability in household behaviour. This cluster set then presents a library of dominant daily consumption patterns that can be used to generate customer archetypes and analyse variability in national residential electricity demand in South Africa. 

We build on previous work where we compared and analysed different clustering techniques for generating daily electricity consumption patterns~\cite{Toussaint2019fair} and developed external clustering evaluation measures from competency questions~\cite{Toussaint2020saicsit}. We validate our results in a use case where we use the pattern library to create customer archetypes for South African households. The archetypes generated by the approach are compared against equivalent benchmark archetypes developed by experts. We show that combining internal and external cluster validation measures enables the selection of a cluster set that is useful for our application. In particular, we found competency questions to be a promising technique for eliciting and representing application requirements. Our approach has potential to enable transparent and repeatable cluster selection by data scientists with limited domain knowledge. 

The paper reviews relevant literature in Section 2, and presents the dataset and clustering experiments in Section~3. In Section~4 we outline our approach to elicit competency questions and formalise application requirements to specify the clustering objective. The clustering results are presented in Section~5. In Section~6 we demonstrate how the pattern library can be applied in a use case to create customer archetypes. Finally, we discuss our findings in Section 7 and conclude in Section 8.

%% file: sections/2_literature.tex
\section{Literature Review}
\label{literature}

This section describes previous work on data processing, algorithms and evaluation measures for clustering electricity consumption patterns, based on the systematic analysis and comparison of clustering approaches in 25 load profile clustering studies published before and in 2018, listed in Table~\ref{tab:energy_clustering_literature} in Appendix~\ref{appendix_b}. We discuss internal and external clustering measures used in the domain, present and discuss the limitations of existing work on cluster analysis for constructing customer archetypes, and introduce competency questions as a method to illicit and specify domain knowledge and application requirements.

\subsection{Clustering Residential Load Profiles}

Long term energy end-use models in the electricity sector are based on information about customers and the dynamics of change \cite{Feinberg2006}. Customer behaviour is frequently approximated with load profiles or load curves, which are time-varying electricity consumption patterns. A daily load profile describes the electricity consumption pattern of a household over a 24 hour period. A representative daily load profile (RDLP) characterises the electricity consumption of a customer archetype~\citep{Chicco2002}, e.g. a rural household on a weekday in winter. RDLPs are obtained by aggregating individual household load profiles that share common attributes, such as socio-demographic characteristics and temporal attributes at varying granularity like a year, season, month and daytype.

\begin{table}[h]
\scriptsize
\centering
{\caption{Frequency of use and performance of clustering algorithms for clustering domestic load profiles, based on 25 studies captured in Appendix \ref{appendix_b} Table \ref{tab:energy_clustering_literature}.}\label{tab:energy_clustering_algorithms_abbreviations}}
\begin{tabular}{clcc}
\textbf{Abbreviation} & \textbf{Algorithm} & \textbf{Frequency of use} & \textbf{Best} \\
\noalign{\smallskip}
& k-means & 19 & 4\\
HC & Hierarchical Clustering & 12 & 2\\
SOM & Self-Organising Maps & 7 & 2\\
& kmedoids & 4 & 2\\
MFTL & Modified Follow-The-Leader & 4 & 2\\
& fuzzy k-means & 4 &\\
& SOM+k-means & 3 & 1\\
& AKM+HC & 3 &\\
& fuzzy c-means & 2 &\\
AKM & Adaptive kmeans & 1 & 1\\
WFAKM & Weighted Fuzzy Averages kmeans & 1 & 1\\
IRC & Iterative Refinement Clustering & 1 & 1\\
MKM & Modified kmeans & 1 &\\
& SAX k-means & 1 &\\
& Spherical k-means & 1 &\\
AVQ & Adaptive Vector Quantisation & 1 &\\
& DBSCAN & 1 &\\
FTL & Follow-The-Leader & 1 &\\
GMM & Gaussian Mixture Model & 1 &\\
& Random Forests & 1 &\\
& Voronoi decomposition & 1 &\\
& SOM+HC & 1 &
\end{tabular}
\end{table}

Cluster analysis is an unsupervised machine learning approach that partitions data points into groups of similar data points~\citep{Aggarwal2015}. It is widely used in the electricity domain to generate RDLPs and construct customer archetypes. We performed an extensive literature survey to find the most widely used and most effective algorithms for clustering electricity consumption patterns. Table~\ref{tab:energy_clustering_algorithms_abbreviations} summarises common clustering algorithms, how frequently they have been used to cluster load profiles and how often they were found to be the best algorithm when multiple algorithms were evaluated, based on the 25 studies that we surveyed. 16 studies compared different algorithms, but 4 of them did not identify the algorithm that was best suited for their application. We assume that this was due to challenges with evaluating clustering structures. For our context and application we concluded that variants of kmeans, self-organising maps (SOM) and hierarchical clustering are the most widely used algorithms. We implemented three of the algorithms: kmeans, SOM, and a combination of the two. We selected the Euclidean distance measure, which was used as similarity measure in the majority of studies and is appropriate in our application where sequences have the same length and a one-to-one mapping between data points.

Clustering residential load profiles has been well explored to understand customer behaviour~\cite{Jin2017}, for demand side management~\cite{Yilmaz2019}, and to create customer archetypes for tariff development \cite{Chicco2002a} and small scale renewable generation \cite{Xu2017} in the electricity sector. Many studies from developed countries in the northern hemisphere cluster relatively homogeneous populations, where domain experts expect that the variability in electricity demand is primarily influenced by seasonal and weekday effects, rather than by extreme variance in consumption between households. Splitting the input data along temporal dimensions prior to clustering is thus common in the literature, for example \citet{Cao2013} split data for summer and winter seasons, and \citet{Dang-Ha2017} additionally split data for weekdays and weekends. \citet{Xu2017} cluster highly variable households spread across the United States. They found pre-binning along a consumer demand dimension, first clustering load profiles by overall consumption and then by load shape, an effective method to improve clustering results in this context. We use a similar two-stage pre-binning with integral kmeans to stratify our variable population along a consumer demand dimension before clustering.

\subsection{Cluster Evaluation}
Despite the practical potential of cluster analysis, evaluating clustering structures remains a challenge for applications in the residential electricity sector. Over 18 different validity metrics were used across the studies we reviewed. Most of them are internal measures, but 3 studies used external measures.

\subsubsection{Internal Clustering Validation Measures}
Internal metrics rely only on information in the data to measure the quality of a clustering and typically evaluate clusters for their compactness and distinctness \cite{Liu2010}. The five internal metrics that have been used most frequently in the studies we surveyed are listed in Table \ref{tab:energy_clustering_metrics_abbreviations}. 

\begin{table}[h]
\centering
\scriptsize
\caption{Frequency count of the five most used clustering metrics for clustering domestic load profiles, based on the 25 studies captured in Appendix \ref{appendix_b} Table \ref{tab:energy_clustering_literature}.}
\label{tab:energy_clustering_metrics_abbreviations}
\begin{tabular}{lll}
\textbf{Abbreviation} & \textbf{Clustering index} & \textbf{Frequency count} \\
\noalign{\smallskip}
DBI & Davies-Bouldin Index & 13 \\
CDI & Cluster Dispersion Index & 11 \\
MIA & Mean Index Adequacy & 11 \\
SMI & Similarity Matrix Indicator & 5 \\
SilhI & Silhouette Index & 4 \\
\end{tabular}
\end{table}

\citet{Jin2017} have found that the ranking of experiments can be inconsistent across internal metrics. They also present a trade-off in their ability to capture both the distinctness and compactness of clusters. \citet{Dang-Ha2017} compare a number of metrics including the CDI, MIA and DBI, and have found that these indicators do not penalise the formation of large, noisy clusters sufficiently, while also tending to create unique clusters for outliers. These limitations have been studied extensively in the data mining literature \cite{Liu2010}, where it is well known that internal metrics tend to be biased towards an algorithm, and that outliers affect clustering outcomes \cite{Aggarwal2015}. Furthermore, a single metric on its own is insufficient to adequately measure the performance of clustering algorithms \citep{Bezdek1998}. We selected the Davies-Bouldin Index \cite{Davies1979}, the Mean Index Adequacy described in \citet{Chicco2002} and the Silhouette Index \cite{Han2012} as internal metrics based on their frequency of use in the domain and ease of implementation for the evaluation of a large dataset.

\subsubsection{External Clustering Validation Measures}\label{sec:external_clustering_measures}
External evaluation measures can be used for additional evaluation when ground truth of the true clusters is known \cite{Aggarwal2015}. Entropy is a widely used external clustering evaluation measure which intuitively measures how class labels are distributed across clusters. \citet{Song2008} define two types of entropy: cluster-based cross-entropy measures the consistency of class labels with respect to clusters; and class-based cross-entropy measures clustering consistency with respect to classes. In the electricity domain \citet{Kwac2014} use entropy as a metric for capturing the daily variability in electricity consumption of households. To evaluate the result of segmenting a large number of daily load profiles into interpretable consumption patterns, \citet{Xu2017} use peak overlap, percentage error in overall consumption and entropy as metrics. We use cluster-based cross-entropy, peak overlap and percentage error to specify the clustering objective for our application. Unlike \citet{Kwac2014} and \citet{Xu2017} who use entropy to measure demand variability within a household, we use entropy to evaluate how consistently clusters are used at specific times and for specific consumption groups.

\subsection{From Clusters to Customer Archetypes}
To associate consumption patterns with characteristic household attributes, current approaches use a two stage process that first clusters load profiles, and then classifies the resultant representative load profiles according to the socio-demographic characteristics of the households that use them \cite{Rhodes2014}\cite{Viegas2016}\cite{McLoughlin2015}. \citet{Viegas2016} and \citet{Rhodes2014} apply context filtering and cluster the average seasonal load profiles of households to derive seasonal patterns. They then perform binary regression to classify the seasonal load curves based on survey data. The study by \citet{Rhodes2014} only considered a small population of 103 college-educated households, with the majority of households earning well above the national mean income. The study by \citet{Viegas2016} was much larger and considered 1972 households from the Irish CER dataset. The same dataset was also used by \citet{McLoughlin2015}, who used a different approach to cluster the daily hourly load profiles of 3941 households and derive consumption patterns by averaging the load profiles of cluster members. They then captured the consumption pattern used by every household on every day in a Customer Class Index (CCI), and assigned the most frequently occurring pattern of the CCI to each household. Finally, they used multinomial logistic regression to classify the CCI by household attributes. 

These approaches for creating customer archetypes capture a very coarse grained temporal dynamic, which does not represent the variability in electricity consumption of households across days, weeks and months. \citet{Cao2013}, \citet{Jin2017} and \citet{Yilmaz2019} present approaches that cluster households' individual daily load profiles, and are thus able to observe variability of household electricity consumption across time and across households. \citet{Kwac2014} cluster individual daily load profiles and build a pattern library to characterise individual household consumption as variable or stable, but the study does exclude low consuming households.  While these studies aim to create patterns that can be used to create customer archetypes or identify households with particular attributes, they do not extend the work to characterise the patterns and evaluate the extent to which they are able to achieve this. This is a typical challenge, as socio-demographic survey data is often unavailable and costly to obtain. We draw on the work of \citet{Jin2017} to capture consumption variability and \citet{McLoughlin2015} to characterise the selected clustering structure in our case study application.

\subsection{Specifying Application Requirements for External Validation}
It is commonly known that the performance of clustering algorithms depends on both the data characteristics and the clustering objective \cite{Aghabozorgi2015}. Aligning cluster evaluation measures with the goal of the specific application thus makes intuitive sense, and is important to generate clusters that are useful \cite{Aggarwal2015}. Ultimately, the clustering process must yield a cluster set that is useful for creating customer archetypes which represent distinguishing socio-demographic attributes of households in the country and the temporal variation in energy consumption. The choice and granularity of these properties are informed by the current practises of experts in the energy sector and the diversity in the population. While some external metrics such as peak overlap, percentage error in overall consumption and entropy have been proposed (see section \ref{sec:external_clustering_measures}), visual inspection by domain experts is commonly suggested and applied as an additional validation step \citep{Dang-Ha2017}. Visual inspection has inherent challenges as suggested by \citet{Gogolou2019}, who have found that the assessment of time series similarity is subjective and depends on its visual representation. Moreover, the application requirements and expert knowledge that are used to evaluate and guide cluster set selection may be implicit and qualitative and difficult to specify explicitly. It would be useful to evaluate clustering structures against the qualitative requirements of domain experts that understand the nuances of the population being clustered, even if they are not familiar with the technicalities of clustering. To do this, methods for formalising qualitative expert knowledge are required. 

The ontology engineering community uses competency questions to acquire context-specific requirements and to compare candidate ontologies \cite{Gruninger1995}.  Competency questions can be used to represent a set of problems that characterise microtheories in a rigorous manner, enabling more precise evaluation of different conceptualisations of a domain \cite{fox1994ontologies}.  Brainstorming, expert interviews and consulting established sources of domain knowledge can be used to identify competency questions \cite{denicola2009}. The techniques for developing competency questions and the questions themselves can be formal or informal. Informal competency questions can be expressed in natural language and connect a proposed ontology to its application scenarios, thus providing an informal justification for the ontology \cite{Uschold96ontologies}. In this study we propose to use unstructured expert interviews to derive competency questions and elicit application requirements. The requirements are then used to guide the evaluation of clustering structures based on their ability to represent the variable daily load profiles of households for the purpose of creating customer archetypes.

%% file: sections/3_clustering.tex
\section{Load Profile Clustering}

In this section we compare and analyse different clustering techniques for generating daily electricity consumption patterns in South Africa. We provide an overview of the application case study and the dataset, the experimental setup which includes clustering algorithms, parameters and preprocessing, and internal clustering validation measures.

\subsection{Application Case Study and Dataset}
\label{data}

The Domestic Electrical Load Metering Hourly (DELMH) dataset \citep{delmh2019} is the largest and most comprehensive database of household electricity consumption in South Africa. South Africa’s electricity utility uses this dataset for long term residential electricity planning. It contains 3~295~194 daily load profiles for South African households over a period of 20 years from 1994 to 2014 \cite{deldata}. 

The daily load profile $h$ for a particular household $j$ on a given day $d$ is a 24 element array containing the mean electricity consumption of the household for each hour of the day. For example, the first element, $l_0$, is the household's mean consumption for the first hour of the day, i.e.  00:00:00 - 00:59:59. 
\begin{equation}
    h^{(j)}_d = 
    \begin{bmatrix}
    l_0&l_1&\ldots&l_{23}
    \end{bmatrix}
\end{equation}
$H^{(j)}$ is a two-dimensional array ($n\_days~\times$ 24) that contains all daily load profiles, $h_d^{(j)}$, for all days, $n\_days_j$, where data is available for household $j$.
\begin{equation}
H^{(j)} =  
    \begin{bmatrix}
    h^{(j)}_0\\
    h^{(j)}_1\\
    \vdots\\
    h^{(j)}_{n\_days_j}
    \end{bmatrix}
\end{equation}
$Y$ is a two dimensional array (3~295~194 $\times$ 24) which concatenates $H^{(j)}$ for all 14~945 households observed over the 20 year period. 

\begin{equation}
Y = 
    \begin{bmatrix}
    H^{(1)}\\
    H^{(2)}\\
    \vdots\\
    H^{(14~945)}
    \end{bmatrix}
\end{equation}

Each row, $y_i$, in $Y$ represents one of the 3~295~194 daily load profiles, $h_d^{(j)}$, across all households $j$ in the data set. 
We can then use clustering to find an optimal clustering structure $k$, given the input dataset $Y$. A single cluster $k_x$ is representative of individual daily load profiles that capture similar daily energy consumption behaviour. The centroid of $k_x$ is used to construct the representative daily load profile (RDLP) of $k_x$, which represents the mean daily consumption pattern of all $y_i$ assigned to the cluster. Collectively, the RDLPs of a cluster set represent the dominant daily consumption patterns across all households in the data set and can be used to generate customer archetypes for long term energy modelling applications.

\subsubsection{Zero Consumption Values and Outliers}
\label{zeroconsumption}
A significant percentage of households in our dataset are low income and rural consumers. In South Africa as in many developing countries, energy access is a priority and must be considered alongside energy security. Through conversations with experts we gathered that the treatment of outlier and extremely low-valued profiles should be reconsidered when clustering electricity consumers in this context. Very low consumption profiles that are close to zero typically belong to consumers living in rural or informal settings. The reasons for low or no consumption are not necessarily due to technical errors, as is assumed in other studies, but because households cannot afford to buy electricity or they choose a different fuel type. The inclusion of these households is important if energy access is a concern. 

\subsection{Experiment Setup}

An experiment run $n$ takes input array $Y$ to produce cluster set $k^{(n)}$ and assigns each normalised daily load profile $y_i'$ to a cluster $k_x^{(n)}$ . The RDLP of each cluster, $r_x^{(n)}$, is the mean of all de-normalised daily load profiles (i.e. $y_i$) assigned to $k_x^{(n)}$.  
A pattern library $\{r_1^{(n)} ... r_{x_n}^{(n)}\}$ is the set of RDLPs for all clusters in $k^{(n)}$. The objective of the load profile clustering experiments is to select the clustering structure $k^{(n)}$ for $Y$ that produces the most useful pattern library for creating customer archetypes. Given the high variance in our dataset, preprocessing was an important component of the clustering process. Different normalisation and pre-binning algorithms were set up for comparison alongside clustering algorithms. 

\subsubsection{Normalisation}
We compared four normalisation techniques from the literature (Table \ref{tab:normalisation_algorithms}) against a baseline with no normalisation.

\begin{table}[h]
\begin{center}
{\caption{Data normalisation algorithms and descriptions}\label{tab:normalisation_algorithms}}
\small
\begin{tabular}{ll p{0.6\linewidth}}
    \hline\noalign{\smallskip}
    \textbf{Normalisation} & \textbf{Equation} & \textbf{Comments} \\
    \noalign{\smallskip}\hline\noalign{\smallskip}
     Unit norm & $y_i' = \frac{y_i}{|y_i|}$ & Scales input vectors individually to unit norm \\ \noalign{\smallskip}
     De-minning & $y_i' = \frac{{y_i} - {y_i}^{min}}{|{y_i} - {y_i}^{min}|}$ & Subtracts daily minimum demand from each hourly value, then divides each value by deminned daily total; proposed by \citet{Jin2017} \\ \noalign{\smallskip}
     Zero-one & $y_i' = \frac{y_i}{{y_i}^{max}}$ & Scales all values to a range [0, 1]; retains profile shape but is very sensitive to outliers; also known as min-max scaler\\ \noalign{\smallskip}
     SA norm & $y_i' = \frac{y_i}{\frac{1}{24} \times \sum_{t=0}^{23} {y_i}[t]}$ & Normalises all input vectors to mean of 1; retains profile shape but very sensitive to outliers; introduced for comparison, as it is frequently used by South African domain experts 
\end{tabular}
\end{center}
\end{table}

\subsubsection{Pre-binning}
\label{sec:prebinning}
We implemented two different approaches to pre-bin all daily load profiles in $Y$. To pre-bin by average monthly consumption ($AMC$), we selected 8 expert-approved bin ranges based on South African electricity tariff ranges (see Appendix \ref{appendix_b} for ranges). All the daily load profiles $H^{(j)}$ of household $j$ were assigned to one of the 8 bins based on the value of the household's average monthly consumption, $AMC^{(j)}$. Individual household identifiers were removed from $Y$ after pre-binning. $AMC$ for household $j$ over one year is:

\begin{equation}
AMC^{(j)}=\frac{1}{12} \sum_{month=1}^{12} \sum_{d=1}^{month_{end}} \sum_{t=0}^{23} 230 \times h_d^{(j)}[t] \textrm{ kWh}
\end{equation}

\noindent
Pre-binning by integral k-means followed these steps:
\begin{enumerate}
\setlength\itemsep{0em}
    \item For each $h_d^{(j)}$:
    \begin{enumerate}
        \item Normalise with unit norm and construct a 24 element vector of the cumulative sum
        \item Append the daily maximum consumption value $h^{(j)}_{d~peak}$
    \end{enumerate}
    \item Concatenate all vectors constructed in step 1 into an array $Y'$ (dim 3 295 194 x 25)
    \item Cluster $Y'$ into $k=8$ bins (same as bins created for $AMC$) with kmeans
    \item Assign all $y_i$ in $Y$ to bins to replicate the clustering structure of $Y'$ \\i.e. a daily load profile $h_d^{(j)}$ of household $j$ on day $d$ should share cluster membership with the same daily load profiles in $Y$ and $Y'$
\end{enumerate}

\subsubsection{Clustering Algorithms}
Variations of kmeans, self-organising maps (SOM) and a combination of the two algorithms were implemented to cluster $Y$. The kmeans algorithm was initialised with a range of $m$ clusters. The SOM algorithm was initialised as a square map with dimensions $s_i \times s_i$ for $s_i$ in range $s$. Combining SOM and kmeans first creates a $s \times s$ map, which acts as a form of dimensionality reduction on $Y$. For each $s$, kmeans then clusters the map into $m$ clusters. The mapping only makes sense if $s^2$ is greater than $m$. $m$ and $s$ are the algorithm parameters.

\subsubsection{Summary of Clustering Experiments}

Table \ref{tab:experiments} summarises the algorithms, parameters and pre-processing steps for each experiment. Each experiment was executed with all normalisation approaches. Experiments with pre-binning were clustered independently in each bin. $Zeros = True$ indicates that zero consumption values were retained in the input dataset. 

\begin{table}[!ht]
\begin{center}
{\caption{Summary of experiments.}\label{tab:experiments}}
\scriptsize
\begin{tabular}{cllcc}
    \hline\noalign{\smallskip}
    \textbf{Exp.} & \textbf{Algorithm} & \textbf{Parameters} & \textbf{Pre-bin} & \textbf{Zeros} \\
    \noalign{\smallskip}\hline\noalign{\smallskip}
     1 & kmeans & $m\{5, 8, 11, ... 136\}$ & none & True\\
       & SOM & $s\{5, 7, 9, ... 29\}$ & none & True\\
       & SOM+kmeans & $s\{30, 40, ... 90\}, m\{5, 8, 11, ... 136\}$ & none & True\\ \noalign{\smallskip}
     2 & kmeans & $m\{5, 8, 11, ... 136\}$ & none & False \\
       & SOM & $s\{5, 7, 9, ... 29\}$ & none & False \\
       & SOM+kmeans & $s\{30, 40, ... 90\}, m\{5, 8, 11, ... 136\}$ & none & False \\ \noalign{\smallskip}
     3 & kmeans & $m\{2, 3, ... 10\}$ & AMC & True\\
       & SOM & $s\{2, 3, 4, 5\}$ & AMC & True\\
       & SOM+kmeans & $s\{4, 7, 11, ... 20\}, m\{2, 3, ... 10\}$ & AMC & True\\ \noalign{\smallskip}
     4 & kmeans & $m\{2, 3, ... 19\}$ & AMC & True\\
       & SOM+kmeans & $s\{4, 7, 11, ... 20\}, m\{2, 3, ... 19\}$ & AMC & True\\ \noalign{\smallskip}
     5 & kmeans & $m\{2, 3, ... 19\}$ & AMC & False \\ \noalign{\smallskip}
     6 & kmeans & $m\{2, 3, ... 19\}$ & integral kmeans & True\\ \noalign{\smallskip}
     7 & kmeans & $m\{2, 3, ... 19\}$ & integral kmeans & False\\ \noalign{\smallskip}
\end{tabular}
\end{center}
\vspace{-2em}
\end{table}

\subsection{Internal Evaluation Measures and the CI Score}
The Mean Index Adequacy (MIA), Davies-Bouldin Index (DBI) and the Silhouette Index were combined into a Combined Index ($CI$) score so that clustering performance can be evaluated across these internal metrics (see Appendix \ref{appendix_c} for details on the metrics). $CI$ is used as a relative index to enable simultaneous interpretation of multiple metrics. Distances between cluster centroids and cluster members were computed using Euclidean distance. The $CI$ is the weighted average $Ix$ score for all bins and calculated as follows:
\begin{eqnarray} \label{eq:CI}
\scriptsize
    CI = log \Bigg( \mathlarger{\mathlarger{\sum}}_{bin=1}^{bins} \Big( Ix_{bin} \times \dfrac{N_{bin}}{N} \Big) \Bigg)
\end{eqnarray}
where $N_{bin}$ is the number of daily load profiles in a bin (as specified in Section~\ref{sec:prebinning}), and $N$ is the total number of daily load profiles in $Y$. 
\begin{equation} 
    Ix_{bin} = \begin{cases}
      \text{undefined }     & \text{if } DBI_{bin}, MIA_{bin} ~or~ Silhouette~Index_{bin} \leq 0\\\\ 
      \dfrac{DBI_{bin} \times MIA_{bin}}{Silhouette~Index_{bin}}    & \text{otherwise}\\
    \end{cases}
\end{equation}
$Ix_{bin}$ is an interim score that computes the product of the DBI, MIA and inverse Silhouette Index in each bin. $CI$ is the log of the weighted sum of $Ix_{bin}$ across all bins. DBI and MIA measure cluster compactness. Both metrics increase as cluster compactness deteriorates, thus increasing $Ix_{bin}$ and $CI$ if this is the case. The Silhouette Index has a range between \{-1, 1\} and is a measure of cluster distinctness and compactness. The Silhouette Index is close to 1 when clusters are both distinct and compact. The closer the Silhouette Index is to 0, the greater $Ix_{bin}$ and $CI$ become. A lower $CI$ score is desirable and an indication of a better clustering structure. The logarithmic relationship between $Ix_{bin}$ and $CI$ means that $CI$ is negative when $Ix_{bin}$ is between 0 and 1, 0 when $Ix_{bin} = 1$ and greater than 0 otherwise. For experiments with pre-binning, the experiment with the lowest $Ix_{bin}$ score in each bin was selected, as it represents the best clustering structure for that bin. For experiments without pre-binning, $bins = 1$ and $N_{bin} = N$. We weighted the $Ix_{bin}$ of each bin to account for the cluster membership size in that bin.

%% file: sections/4_application_requirements.tex
\section{Formalising Application Requirements}
\label{qualitative_evaluation}

In this section we describe how we elicited and formalised implicit domain knowledge from experts who understand the application objective but have limited technical knowledge about clustering techniques. We formulated competency questions from domain knowledge and operationalised them as external evaluation measures, which we implemented as a cluster scoring matrix to provide a qualitative ranking of cluster sets in terms of the application requirements and clustering objective. 

\subsection{Competency Questions and Clustering Objective}
\label{competency_questions}

We analysed existing standards and conducted unstructured interviews with domain experts to formulate informal competency questions expressed in natural language. The Geo-based Load Forecasting Standard (2012) is used as design standard by South Africa's electricity utility and contains manually constructed load profiles and guiding principles for load forecasting in the country. The competency questions were developed after analysis of this standard and continuous engagement with a panel of five industry experts. There were initial interviews with all experts to elicit the usage requirements. Preliminary competency questions were presented at a workshop with key stakeholders in the community. The final version of the competency questions incorporated the feedback from the stakeholders. The following five competency questions were identified and expressed in natural language: 
\begin{enumerate}
\setlength\itemsep{0em}
    \item Does the load shape deduced from clusters represent expected energy demand?
    \item Do clusters distinguish between low, medium and high demand consumers?
    \item Can clusters represent specific loading conditions for different day types and months?
    \item Can a zero-consumption profile be represented in the cluster set\footnote{Important for energy access in low income contexts (see Section \ref{zeroconsumption})}?
    \item Is the number of households assigned to clusters reasonable, given the sample population?
\end{enumerate}
Based on these questions, we define a good cluster set as having expressive clusters and being usable within the context of the intended application. An expressive cluster must convey specific information related to particular socio-economic and temporal energy consumption behaviour. A usable cluster set must represent energy consumption behaviour that makes sense in relation to the application context, and carry the necessary information to make it pertinent to domain users.  

\subsubsection{Cluster Expressivity}

Current domain knowledge suggests that daily electricity consumption behaviour is strongly influenced by daily routines, seasonal climatic variability and the quantity of electricity consumption (low, medium, high) of a household. Beyond producing patterns that exhibit specific features  typically associated with load profiles (question 1), it is desirable that individual clusters convey specific information about the demand profiles of different types of consumers (question 2), on different days of the week and months (question 3).  Expressivity thus requires firstly that the RDLP that a cluster produces is \textit{representative} of the energy consumption behaviour of the individual daily load profiles that are members of that cluster. Secondly, clusters must be \textit{specific} to known temporal and consumption contexts, e.g. low demand households on Sundays in June. The choice and granularity of the temporal and consumption features for enumerating the different contexts must be aligned with and support the accepted practise in the expert community for categorising and analysing the daily load demand of residential households. 

\subsubsection{Cluster Usability}

The attribute of cluster usability was derived from competency questions 4 and 5. Question 4 is evaluated as being either true or false. Question 5 is calculated as the percentage of clusters whose membership exceeds a threshold value. Moreover, while we anticipate a relatively large number of clusters to represent the large variety of consumers, the following two factors should also be considered:
\begin{enumerate}
\setlength\itemsep{0em}
    \item Fewer clusters typically ease interpretation and are thus preferable to larger numbers of clusters
    \item The maximum number of clusters is limited to 220, based on population diversity and existing expert models which account for 11 socio-demographic groups, 2 seasons, 2 daytypes and 5 climatic zones    
\end{enumerate}

\subsection{External Cluster Evaluation Measures}
\label{qualitative_evaluation_measures}
We now translate the clustering objectives into quantifiable external evaluation measures. For \textit{representative} clusters the \textit{mean demand errors} of the \textit{total and peak consumption} values measure the average deviation between the RDLP and the cluster members' load profiles. The \textit{mean peak coincidence ratio} measures the deviation of the peak usage time between the RDLP and the daily load profiles in the cluster. Together these measures express the extent to which a RDLP is representative of the shape and demand of the cluster's member profiles. To measure the degree to which a cluster maps to a $specific$ context, cluster entropy can be used to establish the information embedded in a cluster and thus its specificity. We calculate \textit{day type} and \textit{monthly entropy} to establish \textit{temporal specificity}, and \textit{total} and \textit{peak daily consumption entropy} to establish \textit{demand specificity}.

\subsubsection{Mean Demand Error}
\label{mean_consumption_error}

The total daily demand $h_{total}$ and peak daily demand $h_{peak}$ of a daily load profile are its sum and maximum value respectively. Thus, the total demand $r_{x~total}^{(n)}$ and peak demand $r_{x~peak}^{(n)}$ of a RDLP $r_x^{(n)}$ are its sum and maximum value. Likewise, $h_{l~total}^{x,n}$ and $h_{l~peak}^{x,n}$ are the total and peak daily demand of the member profiles $h_l^{x,n}$ in $k_x^{(n)}$, with $l$ and index counting through all cluster member profiles.
Four error metrics are used to calculate the mean deviation between a RDLP’s peak and total demand, and that of the member profiles. Mean absolute percentage error (MAPE) and median absolute percentage error (MdAPE) are well known error metrics. The median log accuracy ratio (MdLQ)~\cite{Morley2016} overcomes some of the drawbacks of the absolute percentage errors. The median symmetric accuracy (MdSymA) can be interpreted as a percentage error similar to MAPE, making it more intuitive than MdLQ. The equations for calculating the total demand errors are shown below. Both mean and median values are calculated across all $N_x^{(n)}$ cluster members. Equivalent equations are used to calculate peak demand error.

\paragraph{Absolute Percentage Error}
\begin{align}
\scriptsize
    mape & = 100 \times \frac{1}{N_x^{(n)}} \sum_{l=1}^{l=N_x^{(n)}} {\frac{|h_{l~total}^{x,n}-r_{x~total}^{(n)}|}{h_{l~total}^{x,n}}}
\end{align}
\begin{align}
    mdape & = 100 \times median_l \bigg(\frac{|h_{l~total}^{x,n}-r_{x~total}^{(n)}|}{h_{l~total}^{x,n}}\bigg)
\end{align}

\paragraph{Median Log Accuracy ratio}
\begin{align}
    Q & = \frac{r_{x~total}^{(n)}}{h_{l~total}^{x,n}}\\
    mdlq & = median_l \big(log(Q)\big)
\end{align}


\paragraph{Median Symmetric Accuracy}
\begin{equation}
    mdsyma =100 \times (\exp{(median_l \big(|log(Q)|\big))} - 1)
\end{equation}

\subsubsection{Mean Peak Coincidence Ratio}
\label{mean_peak_coincidence_ratio}
We defined peaks as all those values that are greater than half the maximum daily load profile value $h_{peak}$. Peak coincidence is the count that the time of $h_{l~peak}^{x,n}$ coincides with the time of $r_{x~peak}^{(n)}$, the peak demand of the RDLP. Mean peak coincidence averages peak coincidence for all member profiles of $k_x^{(n)}$. The mean peak coincidence ratio is the fraction of mean peak coincidence over the count of peaks in $r_x^{(n)}$. It has a value between 0 and 1. The magnitude of the peak is not considered in the mean peak coincidence ratio.

\subsubsection{Entropy as a Measure of Cluster Specificity}
\label{entropy_measure_cluster_specificity}

Entropy $S$ is used to quantify the specificity of clusters and is calculated as follows:
\begin{equation}
    S_x(F) = - \sum_{i=1}^n {p(f_i) \log_2(p(f_i))}
\label{eq:entropy}
\end{equation}
$F$ is a feature vector with possible values ${f_1, ..., f_n}$. $p(f_i)$ is the probability that daily load profiles with value $f_i$ are assigned to cluster $k_x$. For day type entropy $S_x(day type)$ expresses the specificity of a cluster with regards to day of the week. Thus $F=day type$ has possible values $f_i = \{Mon, Tues, Wed, Thurs, Fri, Sat, Sun\}$. $p(Sun)$ is the likelihood that daily load profiles that are used on a Sunday are assigned to cluster $k_x$. $F=month$ has possible values $f_i = \{January,..., December\}$ and is used to calculate monthly entropy $S_x(month)$. To calculate peak and total daily demand entropy, we created percentile demand bins. Thus the possible values of feature $F=peak\_demand$ are $f_i = \{1,...,100\}$. $p(60)$ is the likelihood that daily load profiles with peak demand corresponding to that of the 60th peak demand percentile are assigned to cluster $k_x$. The lower the entropy, the more information is embedded in the cluster, the more specific the cluster, the better the cluster. 

\subsection{Cluster Scoring Matrix}
The external evaluation measures operationalise the clustering objectives and competency questions as quantifiable scores that can be used to rank experiments. We ranked experiments by each measure and weighted the ranks by the relative importance that experts assigned to that measure. We then calculated a cumulative score for each experiment by summing its weighted ranks. The lower the total score, the better the cluster set meets our application requirements. Table \ref{tab:cluster_scoring_matrix} summarises the objectives, competency questions, qualitative measures and corresponding weights, which we implemented as a cluster scoring matrix. The total score of an external measure for cluster set $k$ is the mean of the individual measures of all clusters $k_x$ with more than 10490 members\footnote{Threshold selected as a value approximately equal to 5\% of households using a particular cluster for 14 days.}. Clusters with a small member size were excluded when calculating mean measures, as they tend to overestimate the performance of poor clusters. Moreover, cluster scores were weighted by cluster size to account for the overall effect that a particular cluster has on the set. For the mean demand error, experiments were ranked against all four error metrics and the mean rank used in the cluster scoring matrix was calculated across all of them.

\begin{table}[h]
{\caption{Clustering objectives, competency questions and external clustering evaluation measures}\label{tab:cluster_scoring_matrix}}
\begin{center}
\scriptsize
\begin{tabular}{p{0.14\linewidth} p{0.04\linewidth} p{0.25\linewidth} c c}
    \hline\noalign{\smallskip}
    \textbf{Objective} & \textbf{CQ} & \textbf{External measure} & & \textbf{Weight} \\
    \noalign{\smallskip}\hline\noalign{\smallskip}
    usable & 4 & zero-profile representation & & 1\\
    \noalign{\smallskip}   
    & 5 & membership threshold ratio & & 2\\
    \noalign{\smallskip}\hline\noalign{\smallskip}
    expressive & 1 & mean demand error & total & 6\\
    representative & 1 & & peak & 6\\
    & 1 & mean peak coincidence & & 3\\ \noalign{\medskip}
    expressive & 3 & temporal entropy & day type & 4\\ 
    specific & 3 & & monthly & 4\\
      & 2 & demand entropy & total daily & 5\\
      & 2 & & peak daily & 5\\
\end{tabular}
\end{center}
\vspace{-2em}
\end{table}

%% file: sections/5_results.tex
\section{Evaluation of Clustering Results}
\label{qual_results}

Experiment runs were conducted using the parameter values in Table \ref{tab:experiments}. Each run was first evaluated with the CI score. The best runs of the best experiments were then further evaluated with the external evaluation measures in the cluster scoring matrix. We implemented our experiments in python 3.6.5 using k-means algorithms from scikit-learn (0.19.1) and self-organising maps from the SOMOCLU (1.7.5) libraries\footnote{The codebase is available online at https://github.com/wiebket/delarchetypes}.

\subsection{Clustering Validation with Internal Measures}
\label{clustering_results}

The CI scores for all experiments range from 2.282 to 9.627. Lower scores are better. Almost two thirds (65.5\%) of experiments have a score below 4. These experiments have been normalised with unit norm, de-minning or zero-one. The remaining experiments have scores above 5 and have not been normalised, or normalised with SA norm. The top 10 ranked experiment runs based on the CI score are shown in Table \ref{tab:clusters_top10}. The percentage point difference between the scores of the first and tenth experiment is only 3.2\%, making it difficult to conclusively select an experiment that is useful for our application.

\begin{table}[h]
{\caption{Top 10 runs ranked by CI score}\label{tab:clusters_top10}}
\scriptsize
\begin{center}
\begin{tabular}{p{0.04\linewidth}ccccp{0.02\linewidth}ccc}
    \hline\noalign{\smallskip}
    \textbf{Rank} & \textbf{CI} & \textbf{DBI} & \textbf{MIA} & \textbf{Sil.} & \textbf{Exp.}& \textbf{Alg.}& \textbf{m} & \textbf{Norm.} \\
    \noalign{\smallskip}\hline\noalign{\smallskip}
1 & 2.282 & 2.125 & 0.438 & 0.095 & 1 & kmeans & 47 & unit \\
2 & 2.289 & 1.616 & 1.220 & 0.262 & 4 & kmeans & 17 & 0-1 \\
3 & 2.296 & 1.616 & 1.220 & 0.260 & 3 & kmeans & 17 & 0-1 \\
4 & 2.301 & 2.152 & 0.485 & 0.119 & 5 & kmeans & 82 & unit \\
5 & 2.316 & 2.115 & 0.447 & 0.093 & 1 & kmeans & 35 & unit \\
6 & 2.320 & 2.199 & 0.486 & 0.121 & 4 & kmeans & 71 & unit \\
7 & 2.349 & 2.152 & 0.481 & 0.143 & 6 & kmeans & 49 & unit \\
8 & 2.351 & 2.189 & 0.434 & 0.090 & 1 & kmeans & 50 & unit \\
9 & 2.354 & 2.111 & 0.476 & 0.128 & 7 & kmeans & 59 & unit \\
10 & 2.355 & 2.173 & 0.453 & 0.093 & 1 & kmeans & 32 & unit \\
\end{tabular}
\end{center}
\vspace{-2em}
\end{table}

Closer analysis of the results confirms that normalisation significantly impacts clustering results. Almost all of the top experiments have been normalised with unit norm, with the exception of two experiments normalised with zero-one. The effects of pre-binning are less clear. Both pre-binning approaches and runs without pre-binning are represented in the top results. Kmeans is the uncontested best clustering algorithm. Four runs belong to Experiment 1 (kmeans, unit norm), but were initialised with different numbers of clusters ($m=\{32, 35, 47, 50\}$). For both the kmeans and SOM algorithms the batch fit time increases linearly with dimensionality. For SOM+kmeans the SOM is used for dimensionality reduction and the dimensions explored are thus considerably greater. This significantly increases experiment run times, as shown in Table \ref{tab:clusters_runtimes}.

\begin{table}[!ht]
{\caption{Summary of algorithm CI scores and run times}\label{tab:clusters_runtimes}}
\scriptsize
\setlength{\tabcolsep}{3pt}
\begin{center}
\begin{tabular}{ccc}
    \hline\noalign{\smallskip}
    \textbf{Algorithm} & \textbf{Mean CI score} & \textbf{Mean run time (s)} \\
    \noalign{\smallskip}\hline\noalign{\smallskip}
k-means & 2.59 & 44.79 \\ \noalign{\smallskip}
SOM & 4.11 & 39.42 \\ \noalign{\smallskip}
SOM + k-means & 3.17 & 1498.77 \\ \noalign{\smallskip}
\end{tabular}
\end{center}
\vspace{-2em}
\end{table}

\subsection{Clustering Validation with External Measures}

The results after external evaluation with the cluster scoring matrix are presented in Table~\ref{tab:qual_ranked_results} for the top runs of the top experiments. The lower the score, the better. The rank by CI score is shown in the last column for comparison. Despite being ranked 9th by CI score, Experiment~7~(kmeans, unit norm) is now ranked 1st.  Table~\ref{tab:clusters_ranked} shows a detailed view of the cluster scoring matrix, with rankings for individual external measures. The second best run, Experiment 4 (kmeans, unit norm), ranks highly for entropy and demand error measures, but has a poorer peak coincidence ratio. Experiment 5 (kmeans, unit norm) ranks third for most measures. While the top two runs lie only 8 points apart, they comfortably outperform the third best run, which has double the score.

\begin{table}[h]
{\caption{Top runs ranked by the total score obtained through validation with external measures}\label{tab:qual_ranked_results}}
\scriptsize
\begin{center}
\begin{tabular}{ccccccc}
    \hline\noalign{\smallskip}
    \textbf{Rank} & \textbf{Score} & \textbf{Exp.} & \textbf{Norm.} & \textbf{Pre-binning} & \textbf{Zeros} & \textbf{CI rank}\\
    \noalign{\smallskip}\hline\noalign{\smallskip}
1 & 57.0 & 7 & unit & int. kmeans & False & 9\\ 
2 & 65.0 & 4 & unit & AMC & True & 6\\ 
3 & 117.5 & 5 & unit & AMC & False & 4\\ 
4 & 143.5 & 6 & unit & int. kmeans & True & 7\\ 
5 & 150.0 & 1 & unit & none & True & 1\\ 
6 & 205.0 & 4 & 0-1 & AMC & True & 2\\ 
7 & 208.0 & 3 & 0-1 & AMC & True & 3\\ 
\end{tabular}
\end{center}
\vspace{-2em}
\end{table}

\begin{table}[h]
{\caption{Cluster Scoring Matrix that ranks experiments by each external evaluation measure}\label{tab:clusters_ranked}}
\scriptsize
\begin{center}
\begin{tabular}{lp{0.05\linewidth}p{0.05\linewidth}p{0.06\linewidth}p{0.05\linewidth}p{0.06\linewidth}p{0.05\linewidth}p{0.05\linewidth}p{0.05\linewidth}}
    \hline\noalign{\smallskip}
    \textbf{Experiment} & & \textbf{1} & \textbf{3} & \textbf{4} & \textbf{4} & \textbf{5} & \textbf{6} & \textbf{7}\\
    \textbf{Normalisation}& & \textbf{unit} & \textbf{0-1} & \textbf{unit} & \textbf{0-1} & \textbf{unit} & \textbf{unit} & \textbf{unit}\\
    \textbf{Qualitative measures} & \textbf{Weight} & & & & & & & \\
    \noalign{\smallskip}\hline\noalign{\medskip}
\textbf{threshold ratio} & \textbf{2}& 1 & 5 & 3 & 5 & 7 & 4 & 1\\ 
\noalign{\smallskip}
\textbf{peak coincidence ratio} & \textbf{3}& 1 & 7 & 4 & 6 & 2 & 5 & 3\\
\noalign{\smallskip}
\textbf{peak demand error} & \textbf{6}& 5.50 & 5.50 & 2.00 & 5.05 & 4.00 & 3.00 & 1.50\\
\textbf{total demand error} & \textbf{6}& 5.00 & 6.25 & 2.00 & 6.00 & 3.25 & 3.75 & 1.00\\ 
\noalign{\smallskip}
\textbf{peak demand entropy} & \textbf{5}& 5 & 7 & 2 & 6 & 3 & 4 & 1\\ 
\textbf{total demand entropy} & \textbf{5}& 5 & 6 & 1 & 6 & 3 & 4 & 2\\
\noalign{\smallskip}
\textbf{day type entropy} & \textbf{4}& 4 & 6 & 1 & 6 & 3 & 5 & 2\\ 
\textbf{monthly entropy} & \textbf{4}& 4 & 6 & 1 & 6 & 3 & 5 & 2\\
\noalign{\smallskip}\hline\noalign{\smallskip}
\textbf{SCORE} & & \textbf{150.0} & \textbf{214.5}& \textbf{65.0} & \textbf{205.0} & \textbf{117.5} & \textbf{143.5} & \textbf{57.0}\\
\end{tabular}
\end{center}
\vspace{-2em}
\end{table}

\begin{figure}[ht]
\begin{center}
  \includegraphics[width=0.75\linewidth]{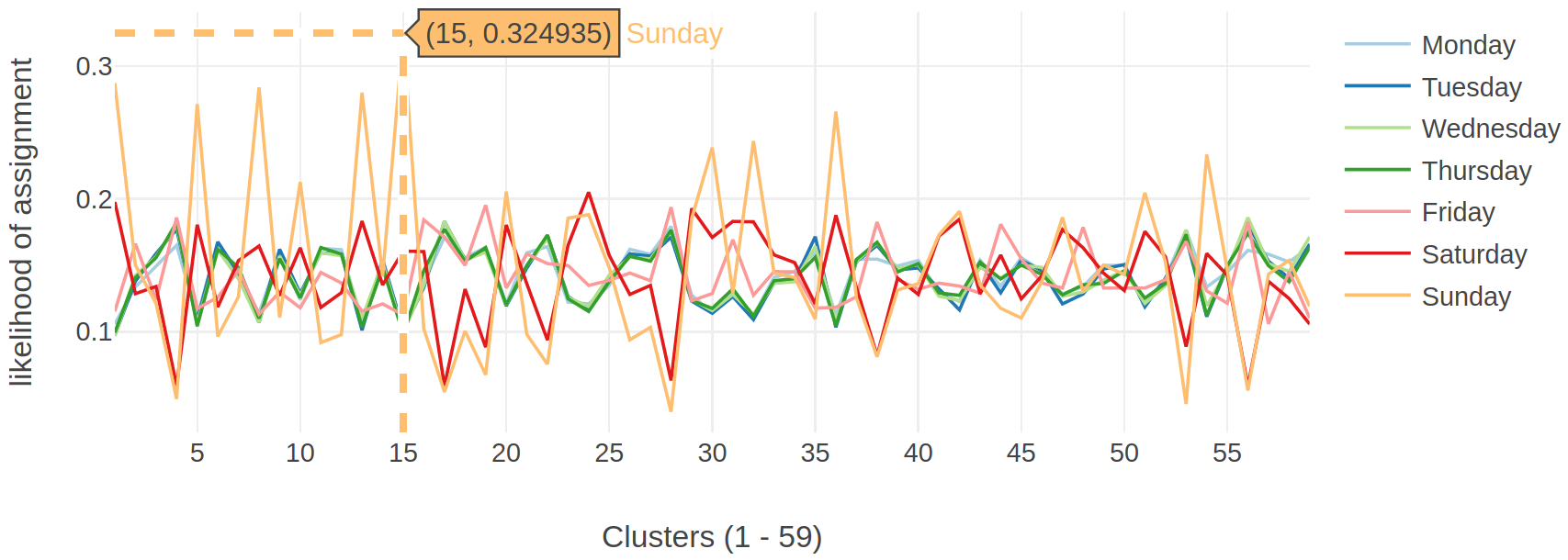}
\caption{Likelihood of day type assignment for Experiment 7 (kmeans, unit norm): the higher (lower) the peak, the more (less) likely that a cluster is used on that day of the week}
\label{fig:temp_homogeneity}      
\end{center}
\vspace{-2em}
\end{figure}

Figure~\ref{fig:temp_homogeneity} visualises the likelihood ($p(f_i)$) that the clusters of Experiment 7 (kmeans, unit norm) are used on a particular day of the week. This is indicative of the entropy (see Eq.~\ref{eq:entropy}) and gives an intuition of the expressivity and usability of the cluster set. The higher the peak of a line, the more likely that profiles assigned to that cluster are used on that day of the week. The lower the peak, the less likely that this is the case. Cluster 15 (C15) is a good example of a cluster that has a very high likelihood of being used on a Sunday, and a lower likelihood of being used on a Saturday or weekday. This cluster is thus specific to the Sunday day type, which is desirable.

\subsection{Contrasting Results of Internal and External Measures}
\label{results_comparison}

\begin{figure}[!ht]
\begin{center}
  \includegraphics[width=0.75\linewidth]{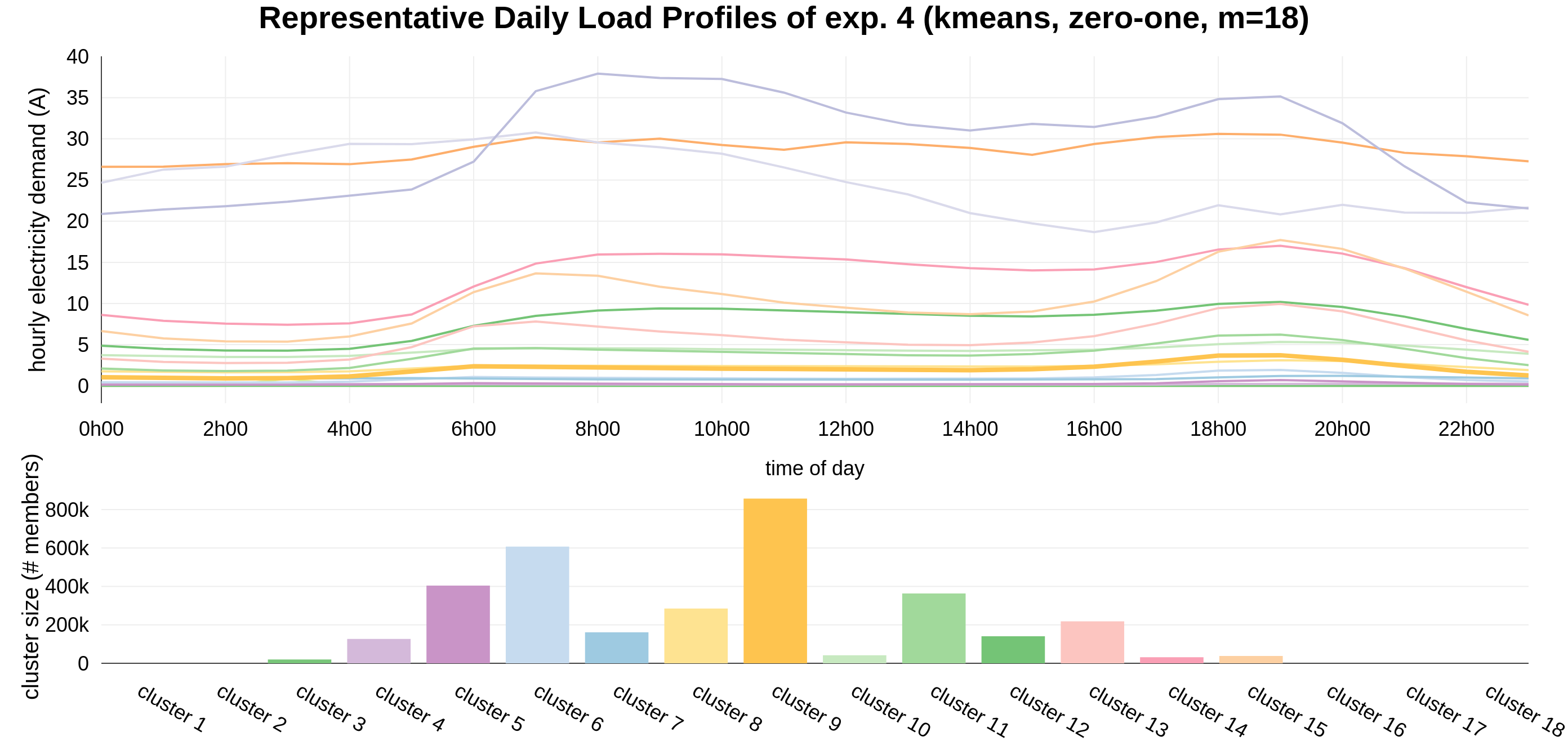}
\caption{RDLPs (top) and cluster membership (bottom) of Experiment 4 (kmeans, zero-one)}
\label{fig:cluster_centroids_exp5_kmeans_zero-one}     
\end{center}
\vspace{-2em}
\end{figure}

\begin{figure}[!ht]
\begin{center}
  \includegraphics[width=0.75\linewidth]{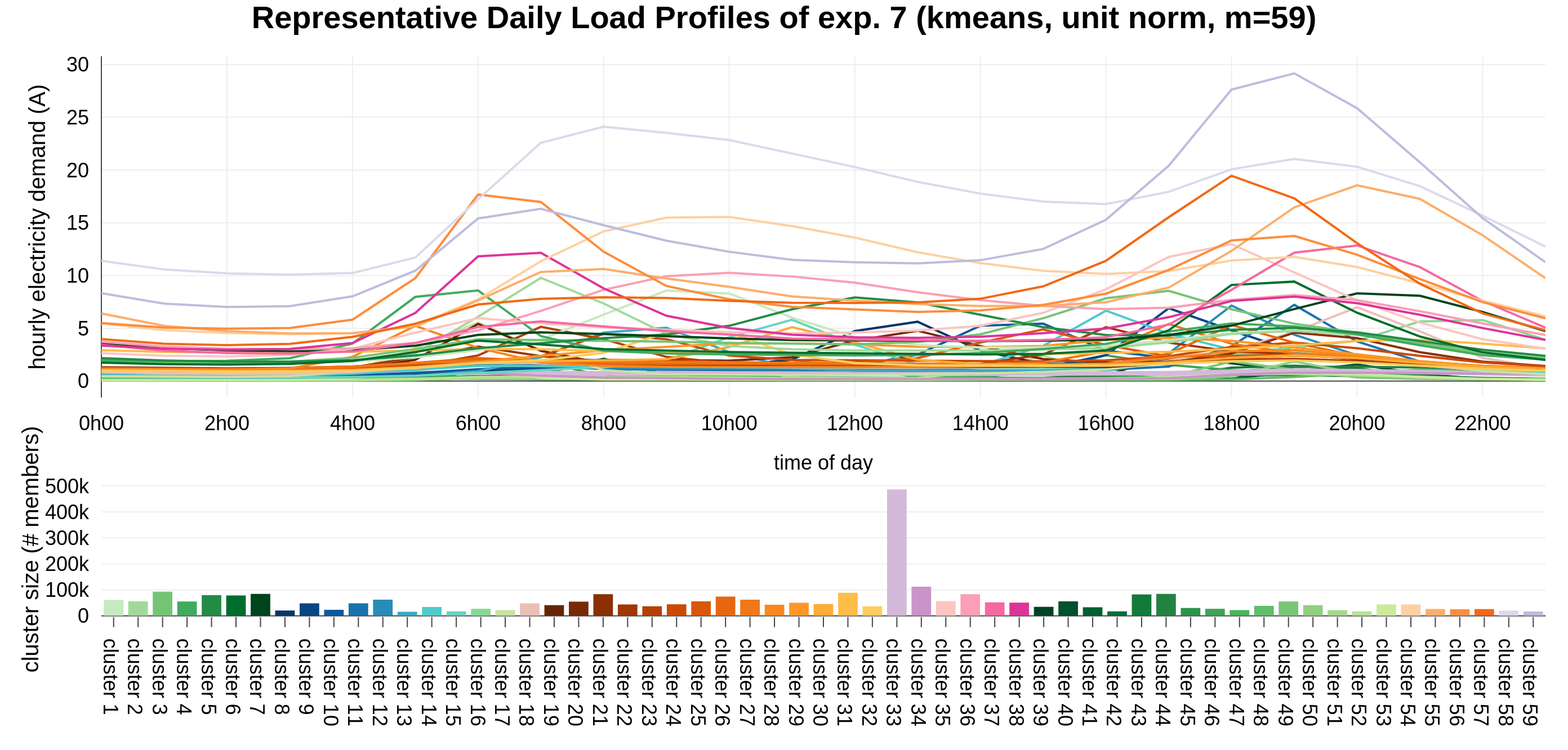}
\caption{RDLPs (top) and cluster membership (bottom) of Experiment 7 (kmeans, unit norm)}
\label{fig:cluster_centroids_exp8_kmeans_unit_norm_bin}
\end{center}
\vspace{-2em}
\end{figure}

The internal metrics provide a useful tool for identifying the most distinct and compact cluster sets, but the CI score is limited for analysis and comparison within the application context. We visually illustrate the strength of the external measures for application-specific evaluation by contrasting the patterns of Experiment 4 (kmeans, zero-one) with those of Experiment 7 (kmeans, unit norm). Experiment 4 (kmeans, zero-one) ranked 2nd based on the CI score, but 6th based on the cluster scoring matrix. Experiment 7 (kmeans, unit norm) on the other hand ranked 9th by CI score, yet ranked 1st after evaluation with external measures. Comparing the patterns in Figures~\ref{fig:cluster_centroids_exp5_kmeans_zero-one} and \ref{fig:cluster_centroids_exp8_kmeans_unit_norm_bin} clearly shows that the latter have greater potential for generating customer archetypes that represent variability in daily electricity consumption behaviour. 

Experiment 4 (kmeans, zero-one) has only 18 clusters. The five smallest clusters combined have fewer than 1500 member profiles and appear invisible in the bar chart at the bottom of Figure~\ref{fig:cluster_centroids_exp5_kmeans_zero-one}. The ragged shapes of the patterns of cluster 16 (C16), C17 and C18 are an indication that very few profiles were aggregated in these RDLPs. Over half of all load profiles belong to only three clusters: C5, C6 and C9. As a whole, the individual RDLPs lack distinguishing features, making them neither expressive nor useable, and thus poor candidates for creating customer archetypes. Experiment 7 (kmeans, unit norm) on the other hand has 59 clusters. With the exception of C33 which accounts for roughly 15\% of all daily load profiles, cluster membership for the remaining clusters varies in a range from 15~000 to 100~000 members. C33 is one of only two clusters in a bin with large membership, due to the high number of low consumption households represented in our dataset. Collectively, the individual patterns are representative and specific, which promises that they will be useful for constructing customer archetypes. 

%% file: sections/6_archetypes.tex
\section{Application of Clusters to Construct Customer Archetypes}
\label{constructing_customer_archetypes}
To validate our selected clustering structure and pattern library for constructing customer archetypes, we benchmark it against customer archetypes currently used by experts in industry. First we describe the benchmark, then we show how our system can be used to create equivalent archetypes. Finally, we illustrate how our pattern library can be used to develop new customer archetypes that capture changing household behaviour. 

\subsection{Benchmark archetype created by experts}
\begin{figure}[!htb]
\centering
	\includegraphics[width=0.8\linewidth]{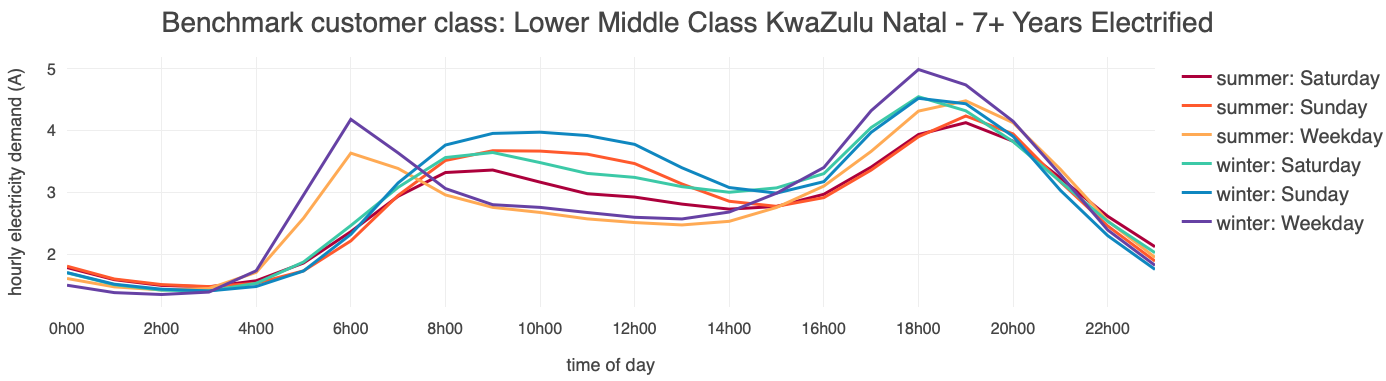}
	\caption{Expert archetype for medium-term electrified lower middle class households in KwaZulu-Natal}
	\label{fig:bm_lowermiddle_KZN_7+} 
\end{figure}

The benchmark customer archetypes represent the aggregated, average electricity consumption of a given type of household, distinguished by its building structure and socio-demographic characteristics. These archetypes are used by South Africa's electricity utility to model the residential load observed at the medium-voltage substation level. We use the archetype of a lower middle class, long term electrified household in KwaZulu-Natal (KZN), South Africa for demonstration purposes. Figure \ref{fig:bm_lowermiddle_KZN_7+} depicts a customer archetype developed by experts for such a household. KZN lies in the East of South Africa, and subsequently has an earlier sunrise and sunset than most other parts of the country. Work day morning peaks are expected between 5am and 7am, and evening peaks between 5pm and 7pm. The climate is subtropical, with humid summers and warm winters. 

Table \ref{tab:expert_archetype_attributes} shows the specific characteristics of different household types identified by experts. The lower middle class household described above will be a household that has piped water access (tap in house), a floor area between 80m$^2$ and 150m$^2$ with walls constructed from asbestos, blocks or bricks and a monthly income between R7 800 and R11 600. 

\begin{table}[htb!]
{\caption{Attributes of South African residential electricity customer archetypes defined by experts}\label{tab:expert_archetype_attributes}}
    \scriptsize
    \centering
    \begin{tabular}{lllll}
    \hline\noalign{\smallskip}
    \textbf{Archetype} & \textbf{Water} & \textbf{Wall material} & \textbf{Floor area} & \textbf{Income} \\
    \noalign{\smallskip}\hline\noalign{\smallskip}
    rural & river/dam & daub/mud/clay & 0-50 & R0-R1.8k \\
    informal & street taps, tap in yard & corr.iron/zinc & 0-50 & R1.8-R3.2k \\
    township & tap in house & asbestos, blocks, brick & 50-80 & R3.2k-R7.8k \\
    lower middle & tap in house & asbestos, blocks, brick & 80-150 & R7.8k-R11.6k \\
    upper middle & tap in house & brick & 150-250 & R19k-R24.5k \\
    \noalign{\smallskip}
    \end{tabular}
\end{table}

\subsection{Archetype reconstructed with pattern library}
We used the clusters from Experiment 7 (k-means, unit norm) to reconstruct the above archetype with a simple multi-class regression model that maps socio-demographic attributes \footnote{The socio-demographic attributes were obtained from the DELSKV \cite{delskv2019} dataset} of cluster members to their clusters. To train the model, we created a feature input vector of the socio-demographic and temporal attributes (i.e. day type and season) for each daily load profile belonging to a cluster. The socio-demographic household data was discretised into the ranges recommended by domain experts as shown in Table~\ref{tab:expert_archetype_attributes}. Each input vector was labelled with the cluster to which the daily load profile was assigned. The model outputs odds ratios that indicate the likelihood that a particular feature value (i.e. socio-demographic or temporal attribute) is correlated with a cluster. We associated attributes with clusters if the odds ratio was equal to or greater than 1.05. The model was trained with WEKA's\footnote{https://www.cs.waikato.ac.nz/ml/weka/} multinomial logistic regression algorithm, but any appropriate classification method can be used. The full implementation details and additional archetypes have been presented in previous work \cite{Toussaint2019msc}.

\begin{figure}[!h]
\centering
\includegraphics[width=0.8\linewidth]{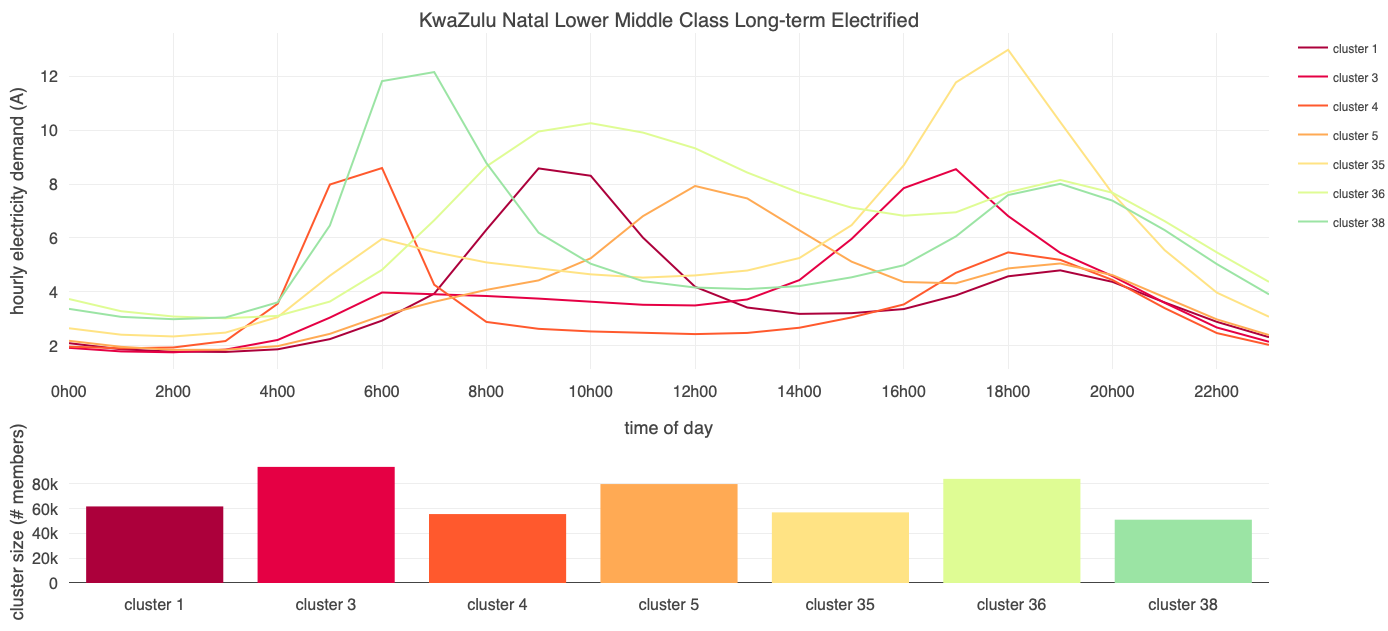}
\caption{RDLPs (top) and cluster membership (bottom) for medium-term electrified lower middle class households in KwaZulu-Natal}
\label{fig:archetype_lowermiddle_KZN_7+}
\end{figure}

Seven clusters showed a strong correlation with the socio-demographic attributes of this archetype. Table \ref{tab:archetype_lowermiddleKZN_rdlp_characteristics} shows the day type and seasonal attributes of the 7 clusters. Each day type in each season is represented by at least one cluster. Full temporal coverage like this is desirable. Work day and weekend clusters, and winter and summer clusters are mutually exclusive. There are 3 winter weekday clusters (C3, C35, C38), one summer weekday cluster (C4), 1 winter weekend cluster (C36) and two summer weekend clusters (C1 and C5). Figure \ref{fig:archetype_lowermiddle_KZN_7+} shows the RDLPs for the clusters. All weekday patterns resemble a typical `out of home' shape, with either a high morning or evening peak and lower consumption throughout the day. This is expected for a lower middle class household, where adults are typically blue collar workers that have a fixed work routine. The patterns of C1, C5 and C36 show a strong correlation with weekends. C1 and C36 are indicative of a slow starting day when there is no job to rush to. C5 with its peak at 12pm is typical for families that have a strong tradition of a shared family lunch on weekends.  The summer weekday pattern of C4 has an earlier morning peak than those of the winter weekday clusters. The weekday patterns of C3, C4, and C35, show an earlier evening peak. With the exception of C3, the winter patterns have a higher energy demand throughout the day than the summer patterns. 

\begin{table}[!htb]
{\caption{Temporal attributes of clusters in Fig \ref{fig:archetype_lowermiddle_KZN_7+}}\label{tab:archetype_lowermiddleKZN_rdlp_characteristics}}
\scriptsize
\centering
\begin{tabular}{c p{0.25\linewidth} c c}
\multicolumn{2}{l}{\textbf{Winter}} & \multicolumn{2}{l}{\textbf{Summer}} \\
\textbf{Cluster} & \textbf{Daytype} & \textbf{Cluster} & \textbf{Daytype} \\
{\bfseries 3} & weekday & {\bfseries 1} & Saturday, Sunday \\
{\bfseries 35} & weekday & {\bfseries 4} & weekday, Friday \\
{\bfseries 36} & Saturday, Sunday & {\bfseries 5} & Saturday, Sunday \\
{\bfseries 38} & weekday, Friday & & \\
\end{tabular}
\end{table}

As a whole, the patterns of this archetype were found to resemble expected customer behaviour. However, some discrepancies exist in relation to the expert archetype. In contrast to the expert RDLPs in Figure \ref{fig:bm_lowermiddle_KZN_7+}, the shapes of our patterns have only one distinct peak, either in the morning or evening. While the peak times correspond between the archetypes, the peak demand values of the expert archetype are approximately half the value of those of our patterns. A plausible explanation for this is that the expert archetype represents the aggregate consumption of a group of households and has only one RDLP for each temporal energy usage context. If we were to aggregate all our patterns for common temporal contexts, for example the three winter weekday RDLPs, the single resultant profile shape and its peak demand would more closely resemble those of the expert archetype.

\subsection{Towards an understanding of dynamic household behaviour}

The method presented in this paper lays the foundation for analysing changing household behaviour and developing new customer archetypes that can inform long term electricity planning and policy interventions. Consider four hypothetical clusters, C1, C2, C3 and C4. After cluster analysis, in a random week, let us assume that household $j$ is assigned to C1 on day $d_1$ and $d_5$, to C2 on $d_2$, $d_3$ and $d_4$, and to C3 on $d_6$ and $d_7$, based on its historical consumption data. A year later, the electricity consumption pattern of household $j$ changes on weekdays, and C4 is now assigned to replace C1 and C2 for $d_1$ to $d_5$. This shift in electricity consumption behaviour is meaningful. Depending on the pattern represented by C1, C2 and C4, it could, for example, be indicative of a high consumption household installing PV panels and reducing grid demand, or of a low consumption household having purchased additional appliances and becoming a medium consumption household. 

The duration and frequency of the shift in consumption is also important and can be used as further evidence to reason about the cause of the change. Do pattern changes happen sporadically, for short periods of time in a season? This may point to extreme weather conditions, like a heat wave. Do they happen consistently at the end of the month? This may be indicative of a household cutting costs due to financial circumstances. Do they coincide with the national lockdown due to Covid-19 that shut down businesses and forced people to stay at home? This could show changes in consumption behaviour that indicate job-loss, or the work-from-home phenomenon. In this manner, external climatic, economic and social circumstances can be linked to the electricity consumption of an individual household using the clusters and pattern library. Exploring this in more detail is an interesting avenue for future work.  

%% file: sections/7_discussion.tex
\section{Discussion}
\label{discussion}

This study combines internal and external validation metrics in the cluster evaluation process to identify a cluster set that represents the daily electricity consumption patterns that best capture variability in behaviour of households in South Africa. As observed in related studies in the electricity domain, we found internal clustering evaluation measures insufficient to capture the nuances and implicit knowledge of domain experts that are needed to identify a useful cluster set. Even though some previous studies have incorporated external validation measures, like entropy, domain knowledge and visual inspection by experts was still required for effective cluster evaluation. Drawing from these studies and the use of competency questions for knowledge elicitation in the ontology engineering community, we operationalised competency questions as external evaluation  measures to identify a cluster set that satisfies the expressivity and usability criteria set by domain experts. 

We conducted unstructured interviews with experts to identify essential characteristics of daily load profiles and customer archetypes for comparing and analysing different cluster sets. The informal nature of unstructured interviews was a good approach for eliciting expert knowledge, as this facilitated an inherently exploratory process through which the pertinent characteristics of RDLPs emerged. We distilled these characteristics into five competency questions for identifying the required expressivity and usability requirements of the application. The competency questions were highly effective for engaging with domain experts, but they lack intrinsic support for specifying the clustering objective. We therefore introduced a collection of external evaluation measures and a cluster scoring matrix to translate the competency questions into a ranking system for evaluating and comparing cluster sets. Unlike previous studies that conducted secondary evaluation steps informally through visual inspection or with evaluation measures that the authors found interesting, the use of competency questions justified our choice of objective external evaluation measures and grounded them in the application requirements. As a whole, the competency questions made assumptions explicit, the external evaluation measures made the clustering objective explicit, and the cluster selection matrix made it easy to apply and repeat the method. By applying the cluster selection matrix during the evaluation step, a data scientist with limited domain knowledge could produce more useful clusters, with limited involvement of domain experts. 

To validate that the clusters selected through our method align with residential load profiles that would be accepted and used by domain experts, we evaluated them in a use case. The use case shows that our clusters can be used successfully to produce customer archetypes that are equivalent to those created by experts. A notable distinction between the RDLPs produced by experts and the RDLPs we derived from our clusters, is that we frequently obtained several patterns for a single day type and season. Each pattern has a distinct shape, peak time and peak demand value. This is a strong indication that our load profiles are more fine grained than the expert ones and better capture the variability in daily consumption of individual households. The profiles of the expert archetypes on the other hand tend to have a morning and evening peak, and lower maximum demand. This is indicative of the aggregate nature of these patterns, which average electricity consumption over a large number of households with different consumption patterns. We observe that our profiles can reconstruct the essential patterns represented in the expert archetypes, which gives us confidence in their usefulness. Additionally, they offer insights about the variability of individual household demand, which opens the door to understanding changing daily consumption behaviour in the residential electricity sector. To our knowledge this paper presents the first end-to-end approach for creating customer archetypes from electricity consumption and survey data in a highly heterogeneous population. The daily consumption pattern library that we have produced provides a mechanism for domain experts to further study the dynamics of household consumption behaviour in relation to household characteristics, like the transition from low to medium electricity consumption, the shift to off-grid renewable generation and potential connections between volatility in electricity use and vulnerability due to social and economic circumstances.

The competency questions, weights and threshold values are subjective, but as our aim was to formalise the domain knowledge that experts use to select clusters and not to create an objective evaluation process, we do not consider this as a constraint. A limitation of the approach however is that eliciting competency questions through unstructured interviews and existing knowledge artifacts requires the synthesis of a large amount of information, which can be tedious and challenging. The initial time investment for creating competency questions and associating them with external evaluation measures is high, and subject to the experts and knowledge sources consulted. However, as utility companies require ongoing insights into customer behaviour on a quarterly or annual basis, this time investment is well spent, as it reduces inconsistencies and the repeated effort required for manually evaluating and selecting clusters. While the audio recordings and notes taken during the informal interviews were sufficient for gathering expert knowledge, it would be worth exploring alternative approaches for eliciting domain knowledge and distilling it into competency questions. Our approach is promising for similar applications in different geographic locations and adjacent domains such as residential water consumption, though the competency questions and external evaluation measures will need to be adapted to suit their objectives.

%% file: sections/8_conclusion.tex
\section{Conclusion}
\label{conclusion}

In this paper we use an application case study to illustrate an approach for eliciting and representing expert domain knowledge and application requirements to formalise clustering objectives and guide the evaluation and selection of clustering structures. We conducted unstructured interviews with experts to identify essential characteristics of daily load profiles and customer archetypes, which we distilled into competency questions. The competency questions were operationalised as external evaluation measures and a cluster scoring matrix. By combining internal and external validation measures we were able to evaluate clustering structures against application requirements to select the clustering structure that best represents the variable daily electricity consumption behaviour of South African households. The selected cluster set was used to create a pattern library and generate customer archetypes that we evaluated against an expert benchmark. Our approach has potential to enable transparent and repeatable cluster ranking and selection by data scientists with limited domain knowledge.

%% file: sections/appendix_b.tex
\section{Load Research Literature}
\label{appendix_b}

Over twenty different algorithms are used for clustering daily load profiles. Many studies have found a particular algorithm to perform better than others. Algorithm abbreviations, their frequency counts and the number of studies that indicate the algorithm as one of the best are listed in Table 
\ref{tab:energy_clustering_algorithms_abbreviations}. Best performing algorithms have been denoted with a * in Table \ref{tab:energy_clustering_literature}. 

\begin{table}[htb!]
\scriptsize
\centering
\begin{tabular}{p{0.17\linewidth} >{\raggedright\arraybackslash}p{0.03\linewidth} | p{0.09\linewidth} p{0.19\linewidth} 
>{\raggedright\arraybackslash}p{0.175\linewidth} >{\raggedright\arraybackslash}p{0.185\linewidth}}
\textbf{Authors} & \textbf{Ref} & \textbf{Input} \newline \textbf{patterns} & \textbf{Customers} & \textbf{Dataset} & \textbf{Notes} \\
\noalign{\smallskip}\hline\noalign{\smallskip}
Batrinu (2005) & \cite{Batrinu2005}& 234 & 234 non-residential & unspecified & IRC developed by authors, builds on Chicco (2003)\\ \noalign{\smallskip}
Bidoki (2010a) & \cite{Bidoki2010a}& 127 & 127 non-residential & unspecified & performance objective dependent\\ \noalign{\smallskip}
Cao (2013) & \cite{Cao2013}& \multicolumn{1}{l}{} & 4225 households & Irish CER dataset & k-means pre-binning \\ \noalign{\smallskip}
Chelmis (2015) & \cite{Chelmis2015}& \multicolumn{1}{l}{} & 115 buildings & USC campus microgrid data & EVI developed by authors\\ \noalign{\smallskip}
Chicco (2002a) & \cite{Chicco2002a} & 471 & 471 non-residential & Romanian national electricity distribution company & \\ \noalign{\smallskip}
Chicco (2003) & \cite{Chicco2003} & \multicolumn{1}{l}{} & 234 non-residential & unspecified  & suggests potential for pre-binning \\ \noalign{\smallskip}
Chicco (2006) & \cite{Chicco2006} & \multicolumn{1}{l}{} & 234 non-residential & unspecified & \\ \noalign{\smallskip}
Dang-Ha (2017) & \cite{Dang-Ha2017} & 3 090 & 3090 households & Hvaler dataset & \\ \noalign{\smallskip}
Dent (2014) & \cite{Dent2014}& 180 & 180 households & NESEMP & clusters \& segments by peak time flexibility \\ \noalign{\smallskip}
Dent (2014a) & \cite{Dent2014a} & 204 & 204 households & NESEMP & assesses variability of energy demand \\ \noalign{\smallskip}
duToit (2016) & \cite{duToit2016} & \multicolumn{1}{l}{} & 11kV \& 22kV feeders & Eskom & shows that PCA dim reduction \& NUBS centroids improve results \& 
run time \\ \noalign{\smallskip}
Figueiredo (2005) & \cite{Figueiredo2005}  & \multicolumn{1}{l}{} & 165 small consumers & Portuguese Distribution Company & \\ \noalign{\smallskip}
Jin (2016) & \cite{Jin2016} & 32 611 421 & residential & unspecified & \\ \noalign{\smallskip}
Jin (2017) & \cite{Jin2017} & 104 673 & 325 households & unspecified & clustering for preprocessing \& data reduction to segment customers \\ 
\noalign{\smallskip}
Kwac (2013) & \cite{Kwac2013} & & 85 households & PG\&E & investigates consistency of consumption \\ \noalign{\smallskip}
Kwac (2014) & \cite{Kwac2014} & 44 949 750 & 123 150 households & PG\&E & AKM developed by authors \\ \noalign{\smallskip}
McLoughlin (2015) & \cite{McLoughlin2015}  & \multicolumn{1}{l}{} & 3941 households & Irish CER dataset & \\ \noalign{\smallskip}
Panapakidis (2018) & \cite{Panapakidis2018} & 365 & 1 small industrial user & unspecified & develops a cluster algorithm selection framework\\ 
\noalign{\smallskip}
Ramos (2012) & \cite{Ramos2012} & & 208 non-residential & Portuguese Distribution Company & \\ \noalign{\smallskip}
Rasanen (2010) & \cite{Rasanen2010} & 3989 & 3989 small consumers & unspecified & \\ \noalign{\smallskip}
Rhodes (2014) & \cite{Rhodes2014} & 103 & 103 households & Pecan Street Project & \\ \noalign{\smallskip}
Teeraratkul (2018) & \cite{Teeraratkul2018} & 23 254 & 1057 households & Opower Corporation (Oracle) & \\ \noalign{\smallskip}
Tsekouras (2007a) & \cite{Tsekouras2007a} & & 94 non-residential & Greek Public Power Cooperation & applies two-stage clustering \\ \noalign{\smallskip}
Viegas (2015) & \cite{Viegas2015} & \multicolumn{1}{l}{} & 1972 households & Irish CER dataset & \\ \noalign{\smallskip}
Xu (2017) & \cite{Xu2017} & 19 070 & residential & Pecan Street Project & best results when applying two-stage clustering\\ \noalign{\smallskip}
\end{tabular}
\caption{Overview of literature on clustering electricity consumers}
\label{tab:energy_clustering_literature}
\end{table}

\begin{table}[htb!]
\scriptsize
\centering
\begin{tabular}{p{0.12\linewidth} >{\raggedright\arraybackslash}p{0.03\linewidth} | >{\raggedright\arraybackslash}p{0.165\linewidth} p{0.09\linewidth} 
p{0.08\linewidth} p{0.08\linewidth} | >{\raggedright\arraybackslash}p{0.1\linewidth} >{\raggedright\arraybackslash}p{0.105\linewidth}} 
 & & \multicolumn{4}{c}{\textbf{Data representation}} & \multicolumn{2}{c}{\textbf{Data variability}} \\
 \textbf{Authors} & \textbf{Ref} & \textbf{RDLP aggregation} & \textbf{Dimensions} & \textbf{Interval} & \textbf{Norm} & \textbf{Time range} & \textbf{Spatial 
cover} \\
\noalign{\smallskip}\hline\noalign{\smallskip}
Batrinu (2005) & \cite{Batrinu2005} & weekday (spring) & 96 & 15min & {[}0,1] & & \\ \noalign{\smallskip}
Bidoki (2010a) & \cite{Bidoki2010a} & annual & 96 & 15min & {[}0,1] & 1 year & \\ \noalign{\smallskip}
Cao (2013) & \cite{Cao2013} & weekday & 48, 18 & 30min, features & {[}0,1] & 4 weeks & Ireland \\ \noalign{\smallskip}
Chelmis (2015) & \cite{Chelmis2015} & none & 96 & 15min & & spring semester Monday & University of Southern California, US \\ \noalign{\smallskip}
Chicco (2002a) & \cite{Chicco2002a} & weekday (winter) & 4 & features & {[}0,1] & 3 weeks & \\ \noalign{\smallskip}
Chicco (2003) & \cite{Chicco2003} & weekday (spring) & 96 & 15min & {[}0,1] & & \\ \noalign{\smallskip}
Chicco (2006) & \cite{Chicco2006} & weekday (spring) & 2 - 6 & features & & & \\ \noalign{\smallskip}
Dang-Ha (2017) & \cite{Dang-Ha2017} & summer, winter, weekday, weekend & 96 & 60min & {[}0,1] & 1 year & Hvaler, Norway \\ \noalign{\smallskip}
Dent (2014) & \cite{Dent2014} & weekday evening peak & 2 - 8 & features & {[}0,1] & & North East Scottland \\ \noalign{\smallskip}
Dent (2014a) & \cite{Dent2014a} & spring weekday evening peak & 48, 42 & 5min, motifs & {[}0,1] & 3 months & North East Scottland \\ \noalign{\smallskip}
duToit (2016) & \cite{duToit2016} & none & 48, 8 & 30min, features & standardise & 2 summer months x 8 years & \\ \noalign{\smallskip}
Figueiredo (2005) & \cite{Figueiredo2005} & summer, winter, weekday, weekend & 96 & 15min & {[}0,1] & & \\ \noalign{\smallskip}
Jin (2016) & \cite{Jin2016} & none & 24 & 60min & de-minning & 1 year & California, US \\ \noalign{\smallskip}
Jin (2017) & \cite{Jin2017} & none & 24 & 60min & de-minning & 1 year & California, US \\ \noalign{\smallskip}
Kwac (2013)  & \cite{Kwac2013} & none & 96, 24 & 15min, 60min & unit norm & 3 summer/ autumn months & City in San Francisco Bay Area, US \\ \noalign{\smallskip}
Kwac (2014)  & \cite{Kwac2014} & none & 24 & 60min & unit norm & 13 months & California, US \\ \noalign{\smallskip}
McLoughlin (2015)  & \cite{McLoughlin2015} & none & 24 & 60min & & 6 months & \\ \noalign{\smallskip}
Panapakidis (2018) & \cite{Panapakidis2018} & none & 24 & 60min & {[}0,1] custom & 1 year & \\ \noalign{\smallskip}
Ramos (2012)  & \cite{Ramos2012} & weekday & 96 & 15min & {[}0,1] & 6 months & \\ \noalign{\smallskip}
Rasanen (2010)  & \cite{Rasanen2010} & & 489 & features & & 1 year & Northern Savo, Finland \\ \noalign{\smallskip}
Rhodes (2014) & \cite{Rhodes2014} & summer, autumn, winter spring & 24 & 60min & {[}0,1] & 1 year & Austin, US \\ \noalign{\smallskip}
Teeraratkul (2018) & \cite{Teeraratkul2018} & none & 24 & 60min & unit norm & 22 days & \\ \noalign{\smallskip}
Tsekouras (2007a)  & \cite{Tsekouras2007a} & & 96 & 15min & {[}0,1] custom & 10 months & \\ \noalign{\smallskip}
Viegas (2015)  & \cite{Viegas2015} & summer, autumn, winter spring & 48 & 30min & & 18 months & \\ \noalign{\smallskip}
Xu (2017) & \cite{Xu2017} & none & 96 & 15min & unit norm & 1 month & 18 cities, US \\ \noalign{\smallskip}
\end{tabular}
\end{table}

\begin{table}[htb!]
\scriptsize
\centering
\begin{tabular}{p{0.13\linewidth} >{\raggedright\arraybackslash}p{0.03\linewidth} | >{\raggedright\arraybackslash}p{0.115\linewidth} 
>{\raggedright\arraybackslash}p{0.25\linewidth} >{\raggedright\arraybackslash}p{0.105\linewidth} | >{\raggedright\arraybackslash}p{0.185\linewidth}}
 &  & \multicolumn{3}{c}{\textbf{Clustering}} & \\
  \textbf{Authors} & \textbf{Ref} & \textbf{Distance measure} & \textbf{Algorithms} & \textbf{Cluster range} & \textbf{Evaluation} \\
   & & &  * is best performing & & \\
\noalign{\smallskip}\hline\noalign{\smallskip}
Batrinu (2005) & \cite{Batrinu2005} & Euclidean & *IRC, HC, k-means, fuzzy k-means, MFTL & 2-15 & ScatI, VRC, MIA, CDI  \\ \noalign{\smallskip}
Bidoki (2010a) & \cite{Bidoki2010a} & Euclidean & k-means, *WFAKM, MFTL, SOM, HC & & MIA, CDI \\ \noalign{\smallskip}
Cao (2013) & \cite{Cao2013} & Manhatten, Euclidean, *correlation, cos & HC, *k-means, SOM+k-means & 5- *14 & peak overlap, Hamming distance, Wiener filter 
\\ \noalign{\smallskip}
Chelmis (2015) & \cite{Chelmis2015} & Euclidean, Hausdorff & k-means, HC, k-medoids, Voronoi decomposition & inconclusive & DI, EVI, VRC \\ \noalign{\smallskip}
Chicco (2002a) & \cite{Chicco2002a} & w-Euclidean & FTL & 7, 9 & MIA, CDI \\ \noalign{\smallskip}
Chicco (2003) & \cite{Chicco2003} & Euclidean, w-Euclidean & k-means, *MFTL, SOM, HC(ward), HC(avg), fuzzy k-means & 10-20; inconclusive & MIA, SMI, 
CDI, DBI \\ \noalign{\smallskip}
Chicco (2006) & \cite{Chicco2006} & Euclidean, w-Euclidean & k-means, *MFTL, SOM, HC(comp), HC(ward), *HC(avg), fuzzy k-means & 10-30; inconclusive & 
CDI, DBI, MDI, ScatI \\ \noalign{\smallskip}
Dang-Ha (2017) & \cite{Dang-Ha2017} & Euclidean, cos, Minkowski & k-means, SKM, *SOM, HC (ward), HC(avg), HC(single) & 2-50; inconclusive & CDI, DBI, MDI, 
MIA \\ \noalign{\smallskip}
Dent (2014) & \cite{Dent2014} & & k-means & 2-10 & CDI, DBI, MIA, SMI, BH \\ \noalign{\smallskip}
Dent (2014a) & \cite{Dent2014a} & Euclidean & k-means, fuzzy c-means, SOM, HC, Random Forests & 8 & MIA, CDI \\ \noalign{\smallskip}
duToit (2016) & \cite{duToit2016} & Euclidean$^2$, DTW, PCC, cos & k-means & *5 & DBI, SilhI \\ \noalign{\smallskip}
Figueiredo (2005) & \cite{Figueiredo2005} & & SOM+k-means & 6-9 & MIA \\ \noalign{\smallskip}
Jin (2016) & \cite{Jin2016} & & AKM+HC & & DBI \\ \noalign{\smallskip}
Jin (2017) & \cite{Jin2017} & Chebyshev, Euclidean & *k-means, *kmedoids, *AKM, *HC(ward), HC(avg), HC(comp), GMM, DBSCAN & 10-500 & CDI, DBI, MIA, SilhI, SMI, 
VRSE \\ \noalign{\smallskip}
Kwac (2013) & \cite{Kwac2013} & & AKM+HC & & threshold \\ \noalign{\smallskip}
Kwac (2014) & \cite{Kwac2014} & & AKM+HC & & threshold, entropy \\ \noalign{\smallskip}
McLoughlin (2015) & \cite{McLoughlin2015} & & k-means, kmedoid, *SOM & 2-16 & DBI \\ \noalign{\smallskip}
Panapakidis (2018) & \cite{Panapakidis2018} & & k-means, MKM(1), MKM(2), various HC, fuzzy c-means, SOM, others & 2-30 & CDI, DBI, MIA, SMI, ScatI, VRC, others 
\\ \noalign{\smallskip}
Ramos (2012) & \cite{Ramos2012} & Euclidean & *k-means, HC (ward), HC (avg), HC(comp), HC(norm cut) & 2-30 & DBI, DI, SilhI, J, others \\ \noalign{\smallskip}
Rasanen (2010) & \cite{Rasanen2010} & Euclidean & *SOM+k-means, SOM+HC(comp) & 2-30 & DBI, IA \\ \noalign{\smallskip}
Rhodes (2014) & \cite{Rhodes2014} & Euclidean & k-means & *2 & \\ \noalign{\smallskip}
Teeraratkul (2018) & \cite{Teeraratkul2018} & Euclidean, *DTW & k-means, *kmedoids, E\&M & & WCS, WB, WCBCR \\ \noalign{\smallskip}
Tsekouras (2007a) & \cite{Tsekouras2007a} & Euclidean & k-means, AVQ, fuzzy k-means, HC & 5-25 & CDI, MIA, SMI, DBI, WCBCR, J \\ \noalign{\smallskip}
Viegas (2015) & \cite{Viegas2015} & & k-means & *2-7 & DBI, DI, SilhI \\ \noalign{\smallskip}
Xu (2017) & \cite{Xu2017} & Euclidean & *k-means, AKM+HC, SAX k-means & 3 (stage 1), 4 (stage 2) & peak overlap, consumption error, entropy, WCSS  \\ 
\noalign{\smallskip}
\end{tabular}
\end{table}

\clearpage

%% file: sections/appendix_c.tex
\section{Supplementary Tables for Clustering Experiments}
\label{appendix_c}

\subsection{Bin ranges AMC pre-binning}
\begin{table}[!ht]
    \small
    \centering
    \begin{tabular}{lcl}
    \noalign{\smallskip}
    \textbf{bin} & \textbf{AMC} & \textbf{} \\
    \noalign{\smallskip}
    1 & 0 - 1 kWh & no consumption \\
    2 & 2 - 50 kWh & lifeline tariff - free basic electricity \\
    3 & 51 - 150 kWh & \\
    4 & 151 - 400 kWh & \\
    5 & 401 - 600 kWh & \\
    6 & 601 - 1200 kWh & \\
    7 & 1201 - 2500 kWh & \\
    8 & 2501 - 4000 kWh & \\
    \end{tabular}
    \caption{AMC bins based on South African electricity tariffs}
    \label{tab:amd_bin_ranges}
\end{table}

\subsection{Clustering metrics}
The Silhouette Index for an individual pattern $p$ in the dataset is:
\begin{equation}
    silhouette(p) = \frac{distinctness(p) - compactness(p)}{max\{distinctness(p), compactness(p)\}} 
\end{equation}
Compactness is the average distance between $p$ and all other patterns in the same cluster. Distinctness is the average distance between $p$ and all remaining patterns that are not in the same cluster.

The Davies Bouldin Index (DBI) for two clusters is calculated as the ratio of the sum of cluster 
dispersions, and the distance between the two cluster centroids.
\begin{equation}
    DBI(i,j) = \frac{dispersion(i) + dispersion(j)}{distance(i,j)} 
\end{equation}
Cluster dispersion can be calculated using different measures. A simple method for computing it is as the average distance between the centroid of a cluster and each pattern in the cluster. The DBI for the dataset is obtained by averaging the similarity measure of each cluster and its most similar cluster, $DBI(i,j)_{max}$, for all clusters. A small DBI value indicates that cluster dispersions are small and distances between clusters are large, which is desirable. When plotting the DBI against the number of clusters, the optimal number of clusters can be visually identified. It is possible for the DBI to have 
several local minima \cite{Davies1979}.